# Voice Controlled Upper Body Exoskeleton: A Development for Industrial Application

Shivam Tripathy, Rohan Panicker, Shubh Shrey, Rutvik Naik, S S Pachpore

**Abstract**— An exoskeleton is a wearable electromechanical structure that is intended to resemble and allow movements in a manner similar to the human skeletal system. They can be used by both disabled and able people alike to increase physical strength in carrying out tasks that would be otherwise difficult, or as a rehabilitation device to aid in physiotherapeutic activities of a weakened body part. This paper intends to introduce a voice-controlled upper body exoskeleton for industrial applications which can aid workers wearing it by reducing stresses on their arms and shoulders over longer periods and add up to 20kg more strength in lifting applications. The 3D design, calculations and considerations, and load analysis are presented along with brief results of a basic prototype model of the exoskeleton.

——————————— ◆ ———————————

## 1 INTRODUCTION

Exoskeletons can be best described as wearable devices that work in tandem with the user. The main function of an Exoskeleton is to amplify the load lifting capabilities of the user, reinforce or support human body or improve human performance or efficiency. They can be considered as a combination of various technologies such as pneumatics, levers, electric motors and several other innovations in material science and design engineering, thus allowing for smooth movement of the limbs while also providing strength and precision-based assistance. The Exoskeleton supports the user's legs, hands, shoulder, waist, thighs etc. while also providing assistance in lifting objects seamlessly and reducing fatigue on muscles. There are basically two types of Exoskeletons: (a) Powered Exoskeleton and (b) Passive Exoskeleton. The exoskeleton that is powered by a system of pneumatics, electric motors and hydraulics among other technologies is a powered exoskeleton while that which does not use such technologies is considered as a passive exoskeleton. A passive exoskeleton however does provide a similar mechanical advantage to the user as that of a powered exoskeleton. For medical applications, powered exoskeletons have improved the quality of life by enabling system-assisted walking to those who have partially or completely lost the ability to walk. Assistance during surgery by enhancing precision is also another positive impact of exoskeletons. An example of exoskeleton applications in defence is the Lockheed Martin's ONYX suit which aims to support soldiers in performing physically draining tasks such as crossing uneven terrains [1][2]. On the other hand, the usage of passive exoskeletons has been in the automotive industry, with the goal of reducing injuries on workers and also errors caused by fatigue. Logistics is another area where their usage has been prominent. Industrial applications of exoskeletons include providing assistance to the upper body and also support to the lower back region during strenuous physical tasks. Several industrial tasks require heavy loads to be lifted from place to place, or up to a specified height. While this could be achieved using a robot or a vehicle to automatically lift the load, it could often prove to be costly and not feasible for the application considering the various parameters needed to carry out the task. Thus, in many cases, the workers of the industry must manually lift the load. This results in great amounts of stress on the arms and shoulders of the worker [4] and very often causes poor posture which could lead to chronic back problems, joint pains and several other medical conditions.

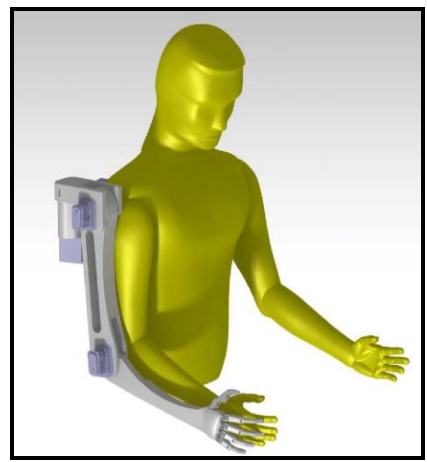

**Figure 1.** FARAT1, a rehabilitation exoskeleton for medical purposes [3]

————————————————
- *Shivam Tripathy - School of Mechanical Engineering, MIT World Peace University, Pune, India. E-mail: shivamtrip99@gmail.com*
- *Rohan Panicker - School of Mechanical Engineering, MIT World Peace University, Pune, India. E-mail: rohanpan27@gmail.com*
- *Shubh Shrey - School of Mechanical Engineering, MIT World Peace University, Pune, India. E-mail: shubhshreypune@gmail.com*
- *Rutvik Naik- School of Mechanical Engineering, MIT World Peace University, Pune, India. E-mail: rutvikn01@gmail.com*
- *S S Pachpore - School of Mechanical Engineering, MIT World Peace University, Pune, India. E-mail: pachporeswanand29@gmail.com*

## 2 DESIGN OF EXOSKELETON

The proposed design is to create a voice controlled - upper-body exoskeleton (to be worn by user) capable of reducing the strain on one's arms and shoulders while lifting the load manually for an extended period of time, and also enabling one to lift up to 20kg more weight than what they would have lifted without wearing the exoskeleton suit. The voice command feature allows the exoskeleton to be easily controlled simply by the voice and through the directions input into a microphone, thus allowing handling of loads even when both hands are occupied in lifting [5].

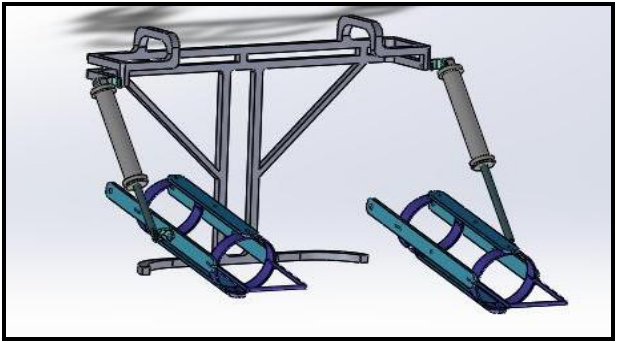

*Figure 2. Computer Aided Design (CAD) of the Exoskeleton*

In order to achieve the desired output, our exoskeleton has three main features working in sync. They are namely Pneumatic System, Frame of the exoskeleton and Electronics System.

### 2.1 Pneumatic System –

The pneumatic system consists of the following OEM parts –
1. Double Acting Pneumatic Cylinder (×2)
2. 5/2 Solenoid Valve (×1)
3. Reservoir (×1)
4. Flow Control Valves (×2)
5. Pneumatic Pipes

The double acting pneumatic cylinders assist in lifting the load and thus allows the user to lift up to 20kg more than what the user could lift without the exoskeleton suit (refer fig-2). These are actuated with the functioning of the 5/2 Solenoid Valve, which integrates with the electronics system and actuates the pneumatic cylinder whenever the voice command is provided. The solenoid valve is provided with compressed air at 6 bar Pressure from the reservoir, which is mounted on the back frame. The rate of actuation of the pneumatic cylinders can be adjusted by the flow control valves.

### 2.2 Frame of the Exoskeleton -

The frame of the exoskeleton consists of the following –
1. Aluminium L-Section Mounts – Manufactured by laser cutting
2. Aluminium Box Sections – As per standard availability
3. Nuts and Bolts - As per standard availability

The frame of the exoskeleton is an integral part as it connects the pneumatic system to the arm of the user. This allows for the efficient movement of the arm of the user about the elbow joint, allowing the user to lift the load to a certain height, restricting the movement in only one plane. The height to be lifted by the user can be easily adjusted by mounting the pneumatic cylinder at different locations of the frame connected to the forearm. The other important feature of the frame is that it ensures that the load being lifted in sync with the strength of the user and the pneumatic piston force applied is transferred throughout the back and thus reducing the loads on the shoulder and arms [6]. The back support of the frame maintains the posture of the user.

### 2.3 Electronics System -

The electronics system consists of the following OEM parts-
1. Arduino UNO Microcontroller (×1)
2. 12V Battery (×1)
3. LM393 sound detection sensor (×1)
4. 5V single channel relay module (×1)
5. Mechanical switch (×1)
6. Electric wires
7. Jumper wires

The electronics system is integral to the exoskeleton as it makes it voice controlled for the convenience and comfort of the user. The sound detection sensor (LM393 op amp) is to be mounted on the exoskeleton such that its' receiver is close to the mouth of the user, to ensure clear reception of voice input. Its sensitivity to sound can be adjusted using the rotary potentiometer in the module. Once the sound sensor detects a sound (voice of the user), it sends a signal to the microcontroller, which passes it on to the relay. The signal pin of the relay is connected to the microcontroller. The relay acts as a switch for operating the solenoid valve. In case of a malfunction of the sound sensor the switch can be used for manual operation or turning the setup off. (refer fig 3) The code uploaded to the microcontroller allows the sound sensor to send signal when it senses a voice and the signal will remain high unless and until it senses the next voice.

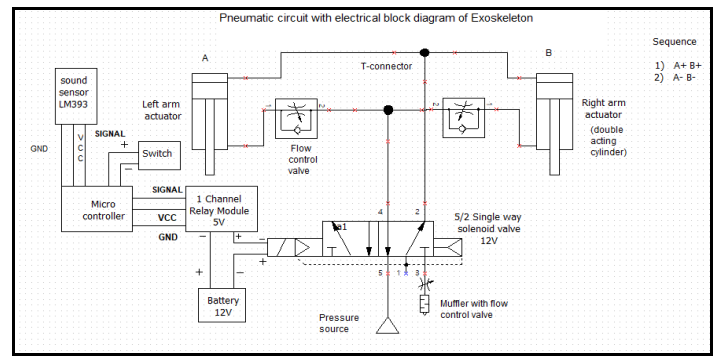

*Figure 3. Complete Circuit of the Exoskeleton*

## 3 ANALYTICAL TREATMENT

### 3.1 Selection of Pneumatic Cylinder

The upper body exoskeleton is designed to apply the equivalent force to lift up to 20kg more weight than without wearing the exoskeleton. That is, the user can lift up to 10kg more weight per arm. In other words, one arm assembly of the exoskeleton consisting of one pneumatic cylinder must provide equivalent force to lift 10kg load. From the side view sketch of one exoskeleton arm, the FBD that can be made is shown in fig.4.

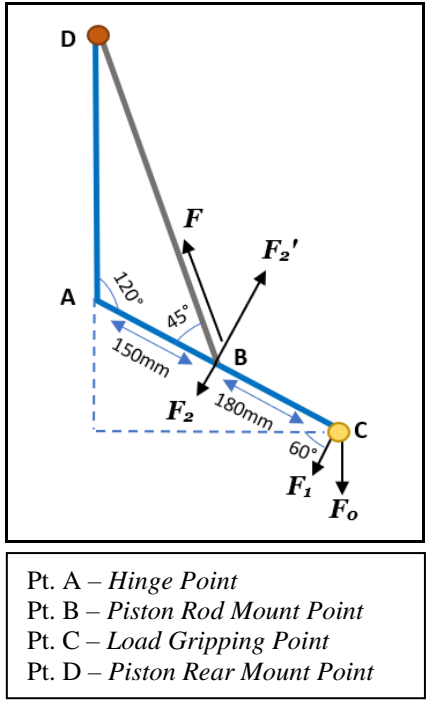

Pt. A – *Hinge Point*
Pt. B – *Piston Rod Mount Point*
Pt. C – *Load Gripping Point*
Pt. D – *Piston Rear Mount Point*

*Figure 4.* FBD of Exoskeleton Arm Arrangement

In Fig. 4, DAC is the transfer angle. While designing a mechanical arrangement consisting of a pneumatic cylinder, transfer angle should not exceed 135 degrees, or else more cylinder force will act against the pivot than be transmitted to lift the load [7]. Hence, the maximum transfer angle at any point in the functioning of the exoskeleton is restricted to 120 degrees, which occurs at the initial position.

### 3.2 Calculation of Force to be Exerted by Piston

From Fig. 4, it can be observed that at Pt. C, a downward force ($F_0$) of 98.1 N is acting, equivalent to the force exerted by the 10kg load. As net moment across AC (forearm) = 0, at start of the lift, the piston force is calculated.
Weight of load $F_0 = 98.1$ N
Weight Perpendicular to AC $F_1 = 84.95$ N
Now the amount of force ($F_2$) exerted at pt. B and perpendicular to AC -

$$\frac{F_1}{F_2} = \frac{AB}{AC} \quad \ldots (1)$$

Upon solving, $F_2 = 186.89$ N
From Fig. 4,
$F_2' = -F_2$ … (2)
So, $F_2 = F_2'$ (in opposite direction)
So, $F_2' = 186.89$ N

Now from Fig. 4,
$F_2'$ is a vector of piston force (F), such that
$$F \times \cos(45) = F_2' \quad \ldots (3)$$
$F = 264.3$ N
Thus, net piston force (F) required to lift a load of 10kg in given mechanical arrangement is 264.3 N.

### 3.3 Calculation of Piston Bore Diameter

$F = P \times A$ … (4)
Where P – Pressure in bar
A – Area in m$^2$
The pressure in the mechanical system is assumed to be a constant pressure (P) of 6 bar;
$P = 6$ bar $= 6 \times 10^5$ Pa
So,
$F = P \times A$
As load will be lifted during retraction [8], the cross-sectional area of piston where compressed air will act –

$$A = \frac{\pi}{4}(D^2 - d^2) \quad \ldots (5)$$

Where D – cylinder bore diameter (m)
d – piston rod diameter (m)
Now, standard piston rod diameter for various applications = 10mm
So, $d = 10^{-2}$ m
From Eq. 1, Eq. 2 and Eq.3,

$$A = \frac{F}{P}$$

$$\frac{\pi}{4}(D^2 - d^2) = \frac{F}{P}$$

Upon solving, D = 0.00256m
Thus, minimum bore of 25.6mm is required for the cylinder to lift 20kg in given mechanical arrangement.
As per standard availability, 30mm is the ideal piston bore diameter required.

### 3.4 Calculation of Piston Stroke Length

The required piston stroke length is calculated using the following constraints –
1. Considering standard arm lengths and measurements of average human male [9].
2. Transfer angle of 120 degrees.

The stroke length achieved which satisfies above constraints is 150mm.
Hence, final specifications of double acting pneumatic cylinder are as follows:
Bore – 30mm
Stroke – 150mm

## 4 VALIDATION OF EXOSKELETON

For validation of design component simulation was performed on simulation platform namely ANSYS Workbench 18.2. The structural analysis of the Upper Body Exoskeleton was done. In the structural analysis, care was taken while applying the boundary conditions so that it perfectly replicates the working environment of our exoskeleton. For structural analysis, output parameters such as Total Deformation, Equivalent Strain and von Mises Stress were taken into consideration.

While performing our analysis following assumptions were made:
1. Material used (i.e. Aluminium; Grade: Al 6061) is isotropic in nature.
2. Exoskeleton Frame is tightly secured to the back portion of the user via means of belt or Velcro Straps.
3. User is using both his hands while lifting the load.

### 4.1 Static Structural Analysis
In order to perform Static Structural Analysis for any model, the STEP/IGES file is imported into the ANSYS Working Environment and then meshing of the model is done and use of course mesh was done for convergence.

#### 4.1.1 Back Support
In the Static Structural Analysis of back support, webs and ribs were considered to be a fixed support. A Force of 98.1N at an angle of 45° was applied at the Pneumatic Cylinder hinge points. Following results were obtained after performing the analysis.

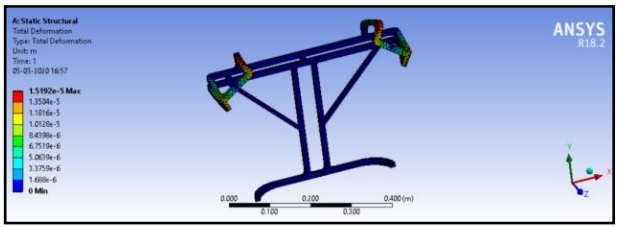

*Figure 5.* Total Deformation

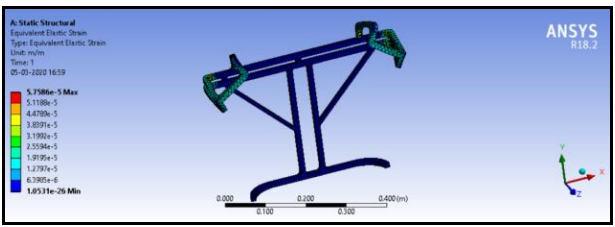

*Figure 6.* Equivalent Strain

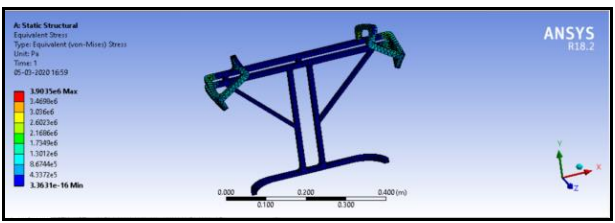

*Figure 7.* Von Mises Stress

The obtained results (Table 1) show that the amount of Total Deformation (Fig 5), Equivalent Strain (Fig 6) and von Mises Stress (Fig 7) produced are not that discouraging and are within the permissible limits. Hence, it is concluded that the Back-Support design is suitable

#### 4.1.2 Wrist Support
In the Static Structural Analysis of wrist support, the end-hinge points were considered to be a fixed support (keeping in mind the locked position of wrist support). A load of 98.1N at an angle of 45° was applied at the rod-end hinge point of the wrist support, which is basically the reaction force exerted by the Pneumatic Cylinder on the section. Additionally, a load of 147.2N (15 kg) was applied in the vertically downward direction at the wrist-end of the wrist support. Reason being the same which was stated in the case of fixed support. Following results were obtained after performing the analysis.

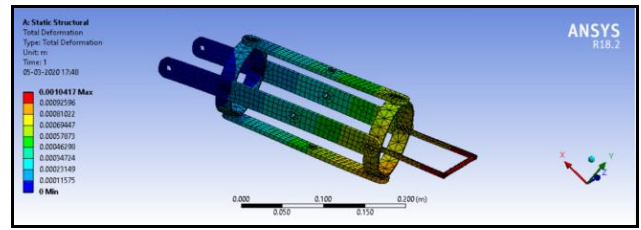

*Figure 8.* Total Deformation

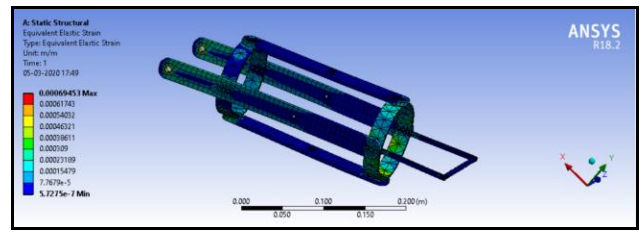

*Figure 9.* Equivalent Strain

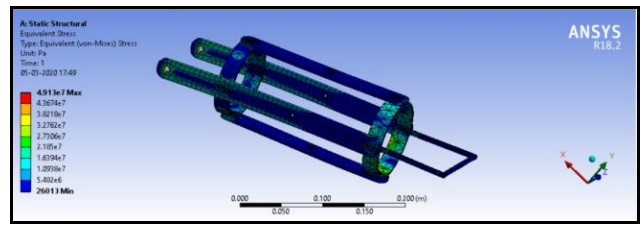

*Figure 10.* Von Mises Stress

The obtained results (Table 1) show that the amount of Total Deformation (Fig 8), Equivalent Strain (Fig 9) and von Mises Stress (Fig 10) produced are not that discouraging and are within the permissible limits. Hence, it is concluded that the Wrist Support design is suitable.

#### 4.1.3 U-Section
In the Static Structural Analysis of U-Section, load of 98.1N at an angle of 45° was applied at both the holes at directions opposite to each other hence, leading to shearing of the section. Back portion of the U-section was considered to be a fixed support while performing the analysis. Following results were obtained after performing the analysis:

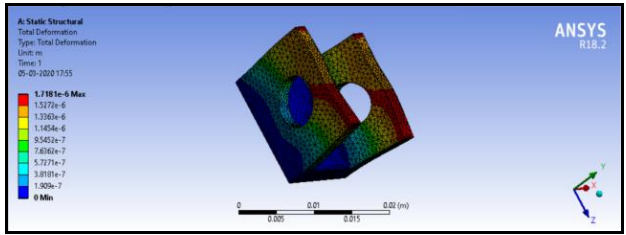

*Figure 11.* Total Deformation

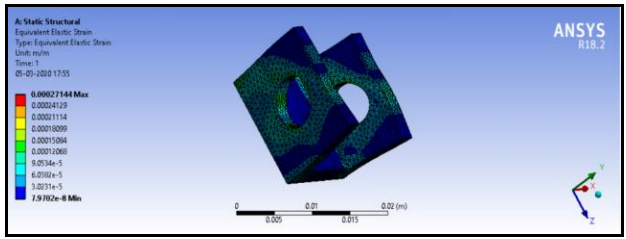

*Figure 12.* Equivalent Strain

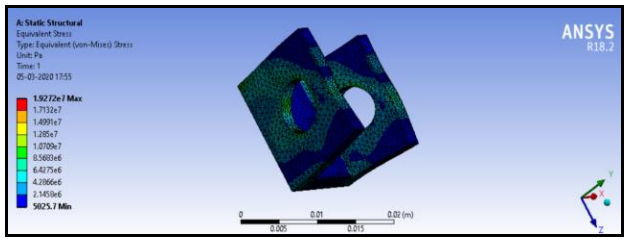

*Figure 13.* Von Mises Stress

The obtained results show that the amount of Total Deformation (Fig 11), Equivalent Strain (Fig 12) and von Mises Stress (Fig 13) produced are not that discouraging and are within the permissible limits. Hence, it is concluded that the U-Section design is suitable.

After performing the Static Structural Analysis for all critical components, following results were obtained as shown in Table 1.

*Table 1.* FEA Result Summary

| Static Structural Analysis | | | | | | |
|---|---|---|---|---|---|---|
| Sr. No | Component Name | Total Deformation (m) | | Equivalent Strain (m\m) | | von Mises Stress (Pa) |
| | | Max | Min | Max | Min | Max | Min |
| 1 | Back Support | 1.52 E-05 | 0 | 5.76E -05 | 1.05E -26 | 3.90E +06 | 3.36 E-16 |
| 2 | Wrist Support | 1.04 E-03 | 0 | 6.95E -04 | 5.73E -07 | 4.91E +07 | 26013 |
| 3 | U-Section | 1.72 E-06 | 0 | 2.71E -04 | 7.97E -08 | 1.93E +07 | 5025 .7 |

After carefully going through the results obtained from Static Structural Analysis, it is concluded that the proposed design is safe in every aspect and in no case, discomfort will be caused to the user. Also, the Factor of Safety (FoS) values obtained from the Stress Analysis of the components fall in the range of 5–15. Further improvements can be made by providing fillet at corners, keeping the cross-section same as far as possible, providing suitable thickness at locations wherever necessary etc.

## 5 PROTOTYPE BUILDING

Once the design was analysed to be safe and feasible, prototype generation was carried out to test the design.

### 5.1 Framework and Pneumatics

Aluminium sheets of thickness 3mm were laser cut and then bent 90 degrees to acquire the desirable mountings of the exoskeleton. The box sections with dimensions 12×12×2mm thick were available in the market. The double acting cylinder of 30mm bore diameter and 150mm was available in the market and was suitable according to the calculations. Initially one arm of the exoskeleton was built. In order to select the ideal position of the double acting cylinder for the arms, testing of the single arm exoskeleton was carried out. The single arm exoskeleton could not transmit the load efficiently to the back and localized the stress on the shoulder. In order to improve the stability of the arm a back support was provided. The back support helps in transmitting part of the load to the lower back, which reduces the load on the shoulder and arms. The mounts used in the back support were riveted. Velcro straps were used as shoulder and back straps for the exoskeleton.

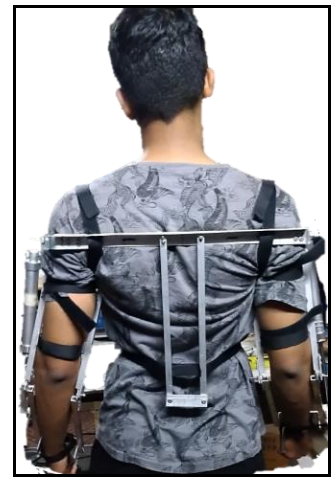

*Figure 14.* Frame of Exoskeleton

### 5.2 Electronics

The initial design consisted of a TIP120 Darlington transistor which had a high time response. This made the actuators prone to sudden movements and jerks, also a breadboard had to be used. This caused loose connections while operating the exoskeleton and making connections every time was tedious. Relay being a mechanical switch has a delay and makes it more comfortable as it would not cause any jerk. Connection of the relay to the battery and the solenoid valve was easy and it was not prone to the loose connections unlike the electronic circuit using the transistor. Hence a single channel 5V relay module was used. Colour coding the wires and a wire harness was used to make the connections understandable.

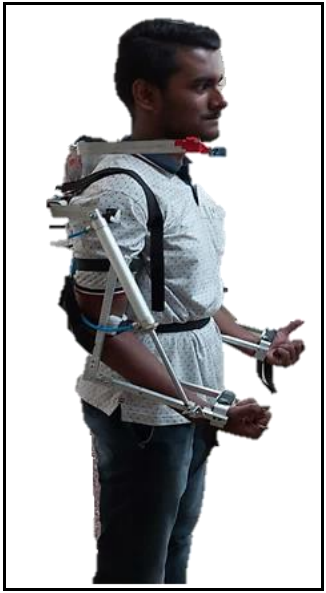

*Figure 15. Basic Prototype of Exoskeleton*

## 6 CONCLUSION

While the exoskeleton proposed in this paper provides great benefits, more complex functionality could be added to the exoskeleton to increase efficiency and allow more tasks to be possible with it. The voice command feature with compact muscle sensors integrated with machine learning can give even better results. Flat brushless DC motors can also be used to increase the number of degrees of freedom of the exoskeleton arms. Pneumatic cylinders can be replaced with pneumatic rotary actuators for specific angular movements. A complex combination of different actuators mounted on the back frame can be employed to transmit the load to the ground by means of legs. Noise cancelling microphones and voice recognition system for the exoskeleton can add further functionality and ease of use. Apart from this the development of new and advanced sensors and improvement in material technologies will help boost the capabilities of exoskeletons in the market.